# Vision-based Automated Bridge Component Recognition Integrated With High-level Scene Understanding

Y. Narazaki[1], V. Hoskere[2], T.A. Hoang[3], B.F. Spencer Jr.[4]

1. Ph.D. student, Department of Civil and Environmental Engineering, University of Illinois at Urbana-Champaign, U.S.A.
   E-mail: narazak2@illinois.edu
2. Ph.D. student, Department of Civil and Environmental Engineering, University of Illinois at Urbana-Champaign, U.S.A.
   E-mail: hoskere2@illinois.edu
3. Ph.D. student, Department of Civil and Environmental Engineering, University of Illinois at Urbana-Champaign, U.S.A.
   E-mail: tuhoang2@illinois.edu
4. Professor, Department of Civil and Environmental Engineering, University of Illinois at Urbana-Champaign, U.S.A..
   E-mail: bfs@illinois.edu

**ABSTRACT**
Image data has a great potential of helping conventional visual inspections of civil engineering structures due to the ease of data acquisition and the advantages in capturing visual information. A variety of techniques have been proposed to detect damages, such as cracks and spalling on a close-up image of a single component (columns and road surfaces etc.). However, these techniques commonly suffer from severe false-positives especially when the image includes multiple components of different structures. To reduce the false-positives and extract reliable information about the structures' conditions, detection and localization of critical structural components are important first steps preceding the damage assessment. This study aims at recognizing bridge structural and non-structural components from images of urban scenes. During the bridge component recognition, every image pixel is classified into one of the five classes (non-bridge, columns, beams and slabs, other structural, other nonstructural) by multi-scale convolutional neural networks (multi-scale CNNs). To reduce false-positives and get consistent labels, the component classifications are integrated with scene understanding by an additional classifier with 10 higher-level scene classes (building, greenery, person, pavement, signs and poles, vehicles, bridges, water, sky, and others). The bridge component recognition integrated with the scene understanding is compared with the naïve approach without scene classification in terms of accuracy, false-positives and consistencies to demonstrate the effectiveness of the integrated approach.

**KEYWORDS:** *Bridge, Viaduct, Columns, Beams, Slabs, Computer vision, Machine learning, object detection, object localization, Multi-scale convolutional neural networks*

## 1. INTRODUCTION

After an earthquake, transportation infrastructure plays a crucial role in supplying necessary resources from the beginning of the initial response. Typically, the post-earthquake response activities have severe time constraints, such as life-saving activities (within 72 hours), supplying critical goods and services (within 72 hours), and providing mass care (within 7 days) [1]. Even when the ground motion at the site does not cause structural collapses, closure of critical transportation for safety inspection and maintenance activities may produce negative social and economic effects, such as people unable to go home [2] and refund of train tickets [3]. To minimize the loss caused by the closure of traffic systems, safety inspection of the transportation facilities need to be carried out rapidly, followed by the retrofit of any observed damage.

Despite the importance of rapid post-earthquake safety inspection of transportation infrastructure, earthquakes often have an adverse effect on the situation in which the inspection is carried out. After the Great East Japan Earthquake, traffic congestion and communication failure hindered the assembly and deployment of human inspectors [4]. Greer [5] mentions that only 20% of the emergency workers were available within seven hours after the occurrence of the 1995 Kobe earthquake. The difficulty of post-earthquake safety inspection by human inspectors suggests that the limited human resources be used strategically, based on the initial estimate of the state of the damage to the transportation infrastructure.

Image data has a potential of assisting the post-earthquake safety inspection by automating the initial damage assessment of the transportation facilities. Image data can be acquired easily and quickly using consumer digital cameras, or potentially using more advanced systems such as satellite imagery [6] and unmanned aerial vehicles (UAVs) [7] [8]. Besides, image data captures the visual information of structures, which is also evaluated during the post-earthquake visual inspection by human inspectors. Therefore, by establishing and implementing appropriate steps of data processing, image data is expected to provide initial knowledge of the current state of the structures without interaction with the human experts.

In civil engineering society, vision-based methods have been frequently applied to damage detection of structural members. For example, detection of cracks on a member surface is attempted using edge maps [9], edge maps with local contextual features [10], and convolutional neural networks (CNNs) [11]. Spalling detection has also been investigated using entropy-based approach [12] etc. However, those damage detection studies assume a close-up image of a single component, and try to distinguish damaged surfaces from healthy surfaces. Therefore, those methods are not guaranteed to work for images containing multiple components of different structures, often raising significant number of false-positive alarms [10].

Research works about damage evaluation of entire structures are relatively sparse and immature. Damage to entire buildings is evaluated using CNNs [13]. However, those approaches are currently limited to the binary evaluation of whether the buildings are damaged or not. Damage evaluation based on three-dimensional information is not reviewed here because the process generally requires significantly more computation than the processing of two-dimensional images, making the process less effective for post-earthquake inspections.

To fill the gap between photos of complex scenes and the component-level vision-based damage evaluation methods, an approach for the automatic detection and localization of critical structural components need to be developed. Although the gap has been recognized and an image-based concrete column recognition method has been proposed [14], the method is based on line segment detection, which does not encode scene-level contextual information. This study focuses on bridges, and develops an approach to bridge component recognition using scene-level contextual information. Bridges are investigated in this study because of their importance in transportation infrastructure, and the visibility of critical structural components, such as column, beams, and slabs. This paper first describes the basic algorithm of the multi-scale convolutional neural networks (multi-scale CNNs), which are implemented to perform pixel-wise classification tasks. Then, the multi-scale CNNs are configured to solve the bridge component recognition task, and the data used during training and testing is explained. Finally, bridge component recognition results are presented and discussed in terms of accuracy, false-positives and consistencies.

## 2. MULTI-SCALE CONVOLUTIONAL NEURAL NETWORKS FOR PIXEL-WISE LABELLING

In this study, bridge components are detected and localized by classifying every pixel of an image into an appropriate category. The classification task is performed by implementing multi-scale convolutional neural networks (multi-scale CNNs), proposed by Farabet et al. [15]. The multi-scale CNN is a compact and effective CNN architecture for the pixel-wise scene labeling, because the network can identify objects photographed in different scales using the same nonlinear filter applied to the images downsampled by different rates (Figure 1(a)).

An image can be represented by a three-dimensional matrix, $\mathbf{I}_0$, whose first two dimensions correspond to the image height and width, and the third dimension indicates the color channel (RGB). By following Farabet et al. [15], the image $\mathbf{I}_0$ is downsampled by 2 and 4 to generate images with smaller sizes, $\mathbf{I}_1, \mathbf{I}_2$, respectively. The collection of downsampled images is called Gaussian pyramid [16]. Note that Farabet et al. [15] used a Laplacian pyramid, which is generated by interpolating lower resolution images and subtracting the interpolated images from the next higher resolution images. However, this study uses the Gaussian pyramid, because many proven network architectures for complex image classification problems are based on the RGB representation, rather than the high-frequency representations like the Laplacian pyramid [17] [18].

The original and the downsampled images are fed into the shared CNN defined by the combination of convolutional layers and maxpooling layers as follows

$$\text{Convolutional layer: } \mathbf{H}_l = f(\mathbf{W}_l * \mathbf{H}_{l-1} + \mathbf{b}_l)$$
$$\text{Maxpooling layer: } \mathbf{H}_l = \text{maxpool}_{N \times N}(\mathbf{H}_{l-1})$$
(1)

In these expressions, $f$ is an activation function, $\mathbf{W}_l$ and $\mathbf{b}_l$ are the convolutional filter and the bias parameters, $N$ is the pooling kernel size, and $\mathbf{H}_l$ is the $l^{th}$ layer output, which is then fed into the $(l+1)^{th}$ layer. The symbol * denotes convolution.

After the series of the convolutional and the pooling layers, the original and the downsampled images, $\mathbf{I}_0$, $\mathbf{I}_1$, $\mathbf{I}_2$, yields the corresponding outputs, $\mathbf{f}_0, \mathbf{f}_1, \mathbf{f}_2$. The multi-scale CNN then upsamples each output into the original scale, and concatenates them to form a multi-scale feature representation of the image, $\mathbf{F}$. This representation is a three-dimensional matrix whose first and second dimension corresponds to the image height and width, and the last dimension corresponds to the features in the three scales.

Finally, the multi-scale feature vector at each pixel location in the image is classified into an appropriate category by placing fully-connected layers (FCLs), which compute the layer output as follows

$$\text{Fully-connected layer (FCL): } \mathbf{H}_l = f(\mathbf{W}_l \mathbf{H}_{l-1} + \mathbf{b}_l)$$
(2)

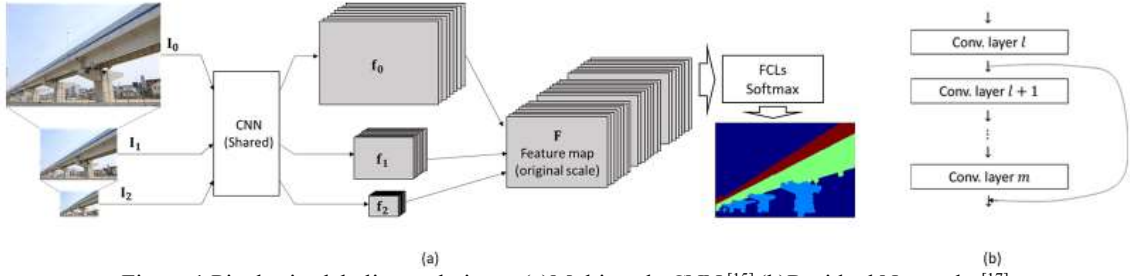

Figure 1 Pixel-wise labeling techniques (a)Multi-scale CNN [15] (b)Residual Networks [17]

Here, $\mathbf{W}_l$ and $\mathbf{b}_l$ are the weight and the bias parameters represented by a two-dimensional matrix and a vector, respectively. As in the convolutional layers, $f$ indicates an activation function, and $\mathbf{H}_l$ is the output of the $l^{th}$ FCL.

The output of the final FCL, $\mathbf{g}$, is a three-dimensional matrix which stores the information of the extent to which each pixel in the image is likely to fall into one of the categories of the classification task. For example, a large value of the $(i,j,k)^{th}$ element, $g_{ijk}$, indicates that the pixel $(i,j)$ is likely to fall into the $k^{th}$ class. The output $\mathbf{g}$ is further transformed into the range $(0,1]$ by placing a softmax layer as follows

$$\text{Softmax layer}: \hat{y}_{ijk} = \frac{e^{g_{ijk}}}{\sum_l e^{g_{ijl}}} \tag{3}$$

where $\hat{y}_{ijk}$ is the $k^{th}$ softmax output at pixel $(i,j)$. The softmax output $\hat{y}_{ijk}$ is close to 1 if the pixel $(i,j)$ is highly likely to belong to the $k^{th}$ category. In contrast, $\hat{y}_{ijk}$ takes a small positive value if the pixel $(i,j)$ is not likely to belong to the $k^{th}$ category. This property allows a probabilistic interpretation of the softmax output. The label maximizing the probability is assigned to each pixel as an estimated label.

The network parameters, $\mathbf{W}_l, \mathbf{b}_l$, are trained by minimizing the cross-entropy loss function between the predicted softmax probabilities and the corresponding target probabilities with an L2-regularization term, or weight decay [19], as follows

$$\text{Cross-entropy loss function with weight decay}: L = -\sum_{ijk} y_{ijk} \ln \hat{y}_{ijk} + \lambda \sum_l \|\mathbf{W}_l\|^2 \tag{4}$$

in which $\lambda$ is a constant parameter, and the weights $\mathbf{W}_l$ include both convolutional filters and FCL weights. The desired probability $y_{ijk}$ is 1.0 if the true label at the pixel $(i,j)$ is $k$, and 0.0 otherwise. In this study, the loss function $L$ is minimized by the Adam optimizer [20].

During the network definition and training, techniques besides the basic operations described above are implemented to improve the quality of the resulting network. Dropout [21] is a technique which randomly samples nodes in a layer with a certain probability during training to reduce overfitting effect. Batch normalization [22] is a method which accelerates the rate of convergence and improves the training results by scaling the layer input to have a new mean and a standard deviation, which are learned during training. Median frequency balancing [23] compensates the data imbalance by weighting the cross-entropy loss in an appropriate manner.

For deep CNNs with many convolutional layers, Residual Networks (ResNets) [17] are effective in improving the training results. The ResNet is characterized by shortcut connections as shown in Figure 1(b), where a connection is established between the $l^{th}$- and the $m^{th}$-layer output. This network models the residual between the desired output of the $m^{th}$-layer and the actual input to the $l^{th}$-layer. All the multi-scale CNN operations including the additional techniques described here is implemented using Python and TensorFlow [24].

## 3. NETWORK ARCHITECTURES

In this study, two multi-scale CNN configurations summarized in Figure 2 are tested to demonstrate the effectiveness of integrating high-level scene understanding with the bridge component recognition. In the first configuration, a multi-scale CNN is trained to estimate the softmax probability corresponding to each of the 10 classes, i.e. Building, Greenery, Person, Pavement, Sign & Poles, Vehicles, Bridges, Water, Sky, and Others. Then, another multi-scale CNN is trained independently to classify the image pixels into five categories, i.e. Non-Bridge, Columns, Beams & Slabs, Other structural, and Other nonstructural. The input to this multi-scale CNN is formed by concatenating RGB image with the softmax probabilities (scaled to [0,255]) corresponding to the first nine categories of the scene classification. The softmax probabilities for the last class are not used explicitly, because the probability is equal to 1 minus sum of probabilities for other classes. Contrary to the two-step classification approach using the scene classifier, the second approach to the bridge component recognition trains a single multi-scale CNN which performs the classification directly from the image.

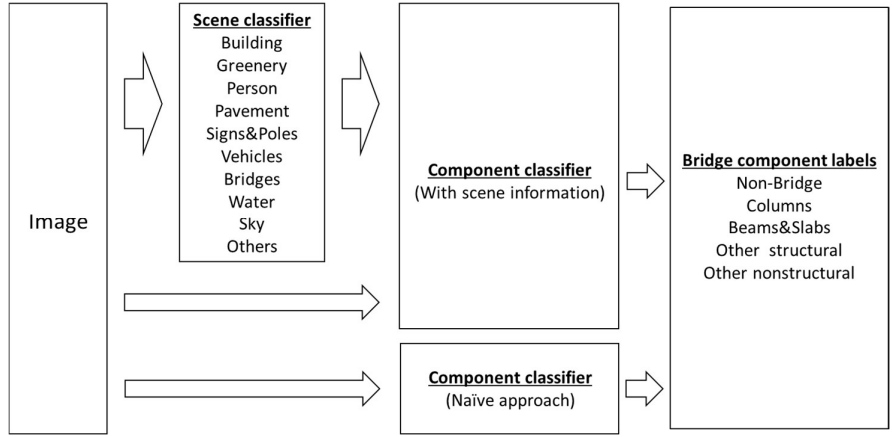

Figure 2 Bridge component recognition approaches with and without scene understanding

For each multi-scale CNN, an appropriate architecture of the convolutional, pooling, and fully-connected layers should be defined. In this study, the architectures shown in Table 1 are tested.

The network *Farabet et al.* is a network used by Farabet et al. [15] to show the effectiveness of the multi-scale CNNs. The main differences of the network in this study from the one in the paper are (i) the use of ReLU activation function instead of the tanh function, (ii) the use of the Gaussian pyramid with RGB image representation instead of the Laplacian pyramid with YUV representation, and (iii) the use of batch normalization and dropout, (iv) optimizer, and (v) the values of training parameters (learning rate, weight decay, batch size).

The network *VGG19_part* is inferred from the VGG19 network [18] for the ImageNet classification problem [25]. The *VGG19_part* network is composed of the first eight convolutional layers of the VGG19 architecture for the shared CNN, and the FCLs with the reduced sizes. The main other differences are (i) the use of batch normalization, (ii) optimizer, and (iii) training parameters.

*ResNet45* implements the shared CNN analogous to the 44-layer residual network used by He et al. [17] to solve the 10-class Cifar-10 classification problem [26]. The main differences are (i) the filter size of the first convolutional layer, (ii) the use of maxpooling instead of the convolution with stride 2, (iii) removal of average pooling after the last convolutional layer, (iv) the addition of one FCL at the end, (v) optimizer, and (vi) training parameters.

*ResNet23* is a multi-scale CNN architecture with 21 shared convolutional layers with ResNet connections and two FCLs. This architecture is deeper than *VGG19_part,* and has larger sizes of filters than *ResNet45*.

The multi-scale CNN architectures described in this section are derived from proven CNN architectures and adapted to the multi-scale representation used in this study. By comparing the performance of these networks and choosing the network with the best performance, the state-of-the-art classifier for this specific problem setting of the bridge component recognition is expected to be obtained.

## 4. DATA COLLECTION AND LABELING

To train the multi-scale CNNs, training data need to be collected for both scene classification and bridge component classification. In this study, the datasets are made by combining the existing datasets and the data manually labelled by the authors.

The training dataset collected for the scene classification is summarized in Table 2. The dataset is subdivided into three categories, General, Urban, and Bridge. General images and labels are collected by combining the Stanford Background Dataset [27] and SIFT Flow dataset [28], both of which are fully-labelled image datasets of broad range of scenes. For the urban scenes, images are acquired from three sources. SYNTHIA dataset [29] provides fully-labelled images of a virtual city viewed from various perspectives and under various lighting conditions. CamVid dataset [30] [31] is a fully-labelled video image sequence of the city of Cambridge. Beside the existing datasets, images of various cities in the USA and Japan are downloaded from the Google Street View [32] using Google Street View Image API [33]. Bridge data consists of bridge images included in the general and the urban datasets, images taken by the authors, bridge and flyover images from the ImageNet dataset [25], Google Street View images, and other images provided by Mr. Shintaro Arai, an amateur photographer. The datasets marked in Table 2 are labelled manually by the authors' research group into 10 categories shown in Table 3.

Before combining the datasets, all images are resized to have 320 pixels in the longer side. Then, the labels of the existing datasets are transferred to those used in this study according to the correspondences shown in Table 3. Finally, night view images and artistic images are excluded, because those images are out of focus of this study.

Table 1 Multi-scale CNN architectures

| Farabet et al. [15] | | ResNet45 [17] | | | ResNet23 | | |
|---|---|---|---|---|---|---|---|
| Name | Filt Size | Name | Filt. Size | ResNet connect. | Name | Filt. Size | ResNet connect. |
| Conv0 | 7x7x16 | Conv0 | 7x7x16 | | Conv0 | 7x7x64 | |
| Maxpool0 | 2x2 | Conv1 | 3x3x16 | | Maxpool0 | 2x2 | |
| Conv1 | 7x7x64 | Conv2 | 3x3x16 | Conv0 | Conv1 | 3x3x64 | |
| Maxpool1 | 2x2 | Conv3 | 3x3x16 | | Conv2 | 3x3x64 | Maxpool0 |
| Conv2 | 7x7x256 | Conv4 | 3x3x16 | Conv2 | Conv3 | 3x3x64 | |
| FCL0 | 1024 | Conv5 | 3x3x16 | | Conv4 | 3x3x64 | Conv2 |
| FCL1 | 10 | Conv6 | 3x3x16 | Conv4 | Conv5 | 3x3x64 | |
| Batch size | 10 | Conv7 | 3x3x16 | | Conv6 | 3x3x64 | Conv4 |
| Wt. decay | 0.0 | Conv8 | 3x3x16 | Conv6 | Conv7 | 3x3x64 | |
| Dropout | 80%@FCL0 | Conv9 | 3x3x16 | | Conv8 | 3x3x64 | Conv6 |
| #param | 1652016 | Conv10 | 3x3x16 | Conv8 | Maxpool1 | 2x2 | |
| | | Conv11 | 3x3x16 | | Conv9 | 3x3x128 | |
| | | Con12 | 3x3x16 | Conv10 | Conv10 | 3x3x128 | Maxpool1 |
| VGG19_part [18] | | Conv13 | 3x3x16 | | Conv11 | 3x3x128 | |
| Name | Filt. Size | Conv14 | 3x3x16 | Conv12 | Conv12 | 3x3x128 | Conv10 |
| Conv0 | 3x3x64 | Maxpool0 | 2x2 | | Conv13 | 3x3x128 | |
| Conv1 | 3x3x64 | Conv15 | 3x3x32 | | Conv14 | 3x3x128 | Conv12 |
| Maxpool0 | 2x2 | Conv16 | 3x3x32 | Maxpool0 | Conv15 | 3x3x128 | |
| Conv2 | 3x3x128 | Conv17 | 3x3x32 | | Conv16 | 3x3x128 | Conv14 |
| Conv3 | 3x3x128 | Conv18 | 3x3x32 | Conv16 | Conv17 | 3x3x128 | |
| Maxpool1 | 2x2 | Conv19 | 3x3x32 | | Conv18 | 3x3x128 | Conv16 |
| Conv4 | 3x3x256 | Conv20 | 3x3x32 | Conv18 | Conv19 | 3x3x128 | |
| Conv5 | 3x3x256 | Conv21 | 3x3x32 | | Conv20 | 3x3x128 | Conv18 |
| Conv6 | 3x3x256 | Conv22 | 3x3x32 | Conv20 | FCL0 | 1024 | |
| Conv7 | 3x3x256 | Conv23 | 3x3x32 | | FCL1 | 10 or 5 | |
| FCL0 | 1024 | Conv24 | 3x3x32 | Conv22 | Batch size | 10 | |
| FCL1 | 1024 | Conv25 | 3x3x32 | | Wt. decay | 0.0001 | |
| FCL2 | 256 | Conv26 | 3x3x32 | Conv24 | Dropout | 80%@FCL0 | |
| FCL3 | 10 | Conv27 | 3x3x32 | | #param | 2403520 (scene) | |
| Batch size | 10 | Conv28 | 3x3x32 | Conv26 | | 2398400 (cpnt,naive) | |
| Wt. decay | 0.0005 | Maxpool1 | 2x2 | | | 2426624 (cpnt,scene) | |
| Dropout | 80%@FCL0 | Conv29 | 3x3x64 | | | | |
| | 80%@FCL1 | Conv30 | 3x3x64 | Maxpool1 | -Wt. decay: weight decay parameter $\lambda$ | | |
| #param | 4423872 | Conv31 | 3x3x64 | | -Dropout: Probability of keeping a node for dropout operation | | |
| | | Conv32 | 3x3x64 | Conv30 | -#param: Number of weight parameters | | |
| | | Conv33 | 3x3x64 | | -cpnt, naïve: Naïve component classifier | | |
| | | Conv34 | 3x3x64 | Conv32 | -cpnt, scene: Component classifier with scene information | | |
| | | Conv35 | 3x3x64 | | | | |
| | | Conv36 | 3x3x64 | Conv34 | | | |
| | | Conv37 | 3x3x64 | | | | |
| | | Conv38 | 3x3x64 | Conv36 | | | |
| | | Conv39 | 3x3x64 | | | | |
| | | Conv40 | 3x3x64 | Conv38 | | | |
| | | Conv41 | 3x3x64 | | | | |
| | | Conv42 | 3x3x64 | Conv40 | | | |
| | | FCL0 | 1024 | | | | |
| | | FCL1 | 10 | | | | |
| | | Batch size | 10 | | | | |
| | | Wt. decay | 0.0001 | | | | |
| | | Dropout | 80%@FCL0 | | | | |
| | | #param | 732464 | | | | |

Table 2 Datasets used for scene classification
(Sources marked with asterisks indicate the images are hand-labelled by the authors)

| Category | Source | # of images | Notes |
|---|---|---|---|
| General | Stanford Background Dataset [27] SIFT Flow dataset [28] | 3403 | 92 bridge photos Night view excluded |
| Urban | SYNTHIA Dataset [29] CamVid Dataset [31] [30] *Google Street View [32] | 6842 | 69 bridge photos Night view excluded |
| Bridge | From General and Urban Datasets | 161 | Fully labelled |
|  | *ImageNet (Bridge, Flyover) [25] *Photos taken by the authors *Others | 1257 | Night view excluded Artistic images excluded Partly labelled |
|  | *Google Street View *Others | 234 | For testing Partly labelled |

Table 3 Label correspondences among datasets used for scene classification
(Labels in existing datasets not listed here are neglected)

| This study | Stanford | SIFT Flow | SYNTHIA | CamVid |
|---|---|---|---|---|
| Building | Building | Awning, Balcony, Building, Door, Staircase, Window | Building, Wall | Archiway, Building, Wall |
| Greenery | Tree, Grass, Mountain | Field, Grass, Mountain, Plant, Tree | Vegetation, Terrain | Tree, VegetationMisc |
| Person |  | Person | Pedestrian, Rider | Bicyclist, Child, Pedestrian |
| Pavement | Road | Crosswalk, Road, Sidewalk | Road, Sidewalk, Parking-slot, Lanemarking | LaneMkgsDriv, LaneMkgsNonDriv, ParkingBlock, Road, RoadShoulder, Sidewalk |
| Sign&Poles |  | Pole, Sign, Streetlight | Pole, Traffic sign, Traffic light | Column_Pole, SignSymbol, TrafficLight |
| Vehicles |  | Bus, Car | Car, Bicycle, Motorcycle, Truck, Bus, Train | Car, MotocycleScooter, OtherMoving, SUVPickupTruck, Train, Truck_Bus |
| Bridges |  | Bridge |  | Bridge |
| Water | Water | River, Sea |  |  |
| Sky | Sky | Moon, Sky, Sun | Sky | Sky |
| Others |  | Bird, Boat, Cow, Fence | Fence | Animal, CarLuggagePram, Fence, TrafficCone |

Table 4 Bridge component classification dataset
(Sources marked with asterisks indicate the images are hand-labelled by the authors)

| Source | # of images | Notes |
|---|---|---|
| *ImageNet (Bridge, Flyover) [25] *Photos taken by the authors *Others | 1135 | Bridge photos Night view excluded Artistic images excluded |
| Google Street View [32] | 194 | Non-bridge photos |
| *Google Street View *Others | 234 | For testing |

The general and urban images are shuffled within each category, and saved as data blocks containing up to 250 images. For the general and the urban images, the first 90% of the images in each data block are used during training, and the remaining 10% are used during testing.

For the bridge dataset, testing data are separated from training data as shown in Table 2 to assure independence between training and testing data. Then, the training images are saved as data blocks containing up to 250 images. The bridge testing images are saved into a single file.

The training data for the bridge component classification consists of bridge and flyover images from ImageNet, Google Street View images, and the images taken by the authors and the amateur photographer, as shown in Table 4. All the images listed in the table are hand-labelled by the authors research group into one of the five classes, i.e.

Non-bridge, Columns (including piers), Beams & Slabs, Other Structural (trusses, arches, cables, abutments, remarkable braces, remarkable bearings etc), and Other Nonstructural. The same 234 testing images for the scene classification are used again for the testing of the bridge component classification. Each of the training and testing dataset is saved as a single data block.

## 5. BRIDGE COMPONENT RECOGNITION RESULTS

This section trains and tests the bridge component classifiers described previously using the collected datasets. The training begins with loading appropriate image data blocks. Then, the training images in each data block are shuffled, augmented, and used to update the network parameters. Following Farabet et al. [15], the image augmentation includes random resizing (between 75% and 125% for general and the urban datasets, between 50% and 150% for bridge dataset), random cropping (180×180 image is cropped from 320×320 image), random rotation up to $\pm 15°$, random flipping, and jitter (zero-mean normal distribution with $\sigma = 2$).

### 5.1 Scene Classification

During the training of the scene classifier, an image data block is loaded from each of the general, urban, and bridge datasets in a random order. After shuffling and augmenting the training images in the data blocks, a mini-batch of 10 images are made by choosing four images from the general category, four images from the urban category, and two images from the bridge category. The creation of the mini-batch and the network parameter update using the mini-batch are repeated until all the training images in one of the loaded data blocks are seen by the network. The iterative process until the network sees all data blocks is called one cycle of training.

The network architectures *Farabet et al.*, *VGG19_part*, *ResNet45*, and *ResNet23* defined in Table 1 are trained for 65 cycles in total, where the learning rate is gradually decreased from $10^{-4}$ (first 50 cycles) to $10^{-5}$ (the next 10 cycles), and $10^{-6}$ (last 5 cycles).

The total pixel-wise accuracies evaluated on the testing data are 80.96% (*Farabet et al.*), 87.48% (*VGG19_part*), 86.70% (*ResNet45*), and 88.68% (*ResNet23*). The confusion matrix of the best network, *ResNet23*, is shown in Figure 3. The confusion matrix shows that all classes except 'Signs&Poles' and 'Others' are recognized with pixel-wise accuracy higher than 80%, including 94% accuracy for the bridge class. Therefore, the *ResNet23* architecture is used to provide the bridge component classifier with the scene information.

### 5.2 Bridge component recognition

The bridge component classifiers with and without scene understanding are then trained using the dataset described in Table 4. After loading, shuffling, and augmenting the training images, a mini-batch of 10 images are made and used to update the network parameters. Because the training images are saved as a single data block in this case, one cycle of training refers to the iterative process until the network sees all the training images stored in the block. The multi-scale CNNs with *ResNet23* architecture is trained for 600 cycles in total. The learning rate is $10^{-4}$ for the first 500 cycles, $10^{-5}$ for the next 180 cycles, and $10^{-6}$ for the last 20 cycles.

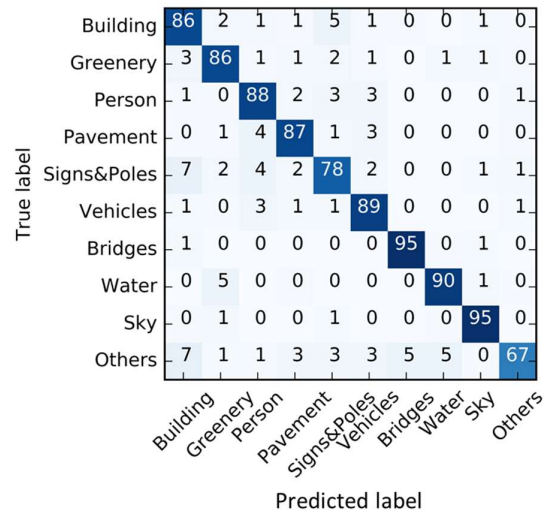

Figure 3 Confusion matrix (Scene classifier, ResNet23)

The total pixel-wise accuracies of the trained classifier with and without scene understanding are 80.34% and 82.41%, respectively. The corresponding confusion matrices are also shown in Figure 4. By integrating the scene understanding, the total accuracy decreases by 2%, which could be explained by the error of the scene classifier.

The example column detection results are shown in Figure 5. For the viaduct and the cable stayed bridge, both classifier could recognize bridge components with minor errors. For the image of arch bridge with another bridge in the background, the both classifier could identify the part of the bridges although the results are less accurate. Significant improvement of the bridge component recognition results by integrating scene understanding can be observed for the non-bridge image. Within the image, the classifier with scene understanding finds almost no bridge component. In contrast, the classifier without scene understanding shows remarkable false positives.

To further examine the effect of integrating scene understanding on the false-positive detections, the bridge component classifiers with and without scene understanding are applied to non-bridge images used during the testing of the scene classification. Because the images do not contain any bridge, the perfect classifier should label all pixels as non-bridge. The results of the false-positive test shown in Table 5 reveal that the bridge component classifier integrated with the scene understanding raises remarkably less false-positive detections than the classifier without scene understanding. Therefore, the bridge component recognition with scene understanding is effective in finding and localizing bridge components from complex scenes with various irrelevant objects.

The example results also show inconsistencies, such as the isolated false positive structural components and the structural members which is not continuous with any other component. Although the inconsistencies by false positive detections are reduced by integrating scene information, the approach proposed in this paper is not enough to eliminate the inconsistencies. To reduce the inconsistencies, appropriate post-processing techniques should be applied in the future.

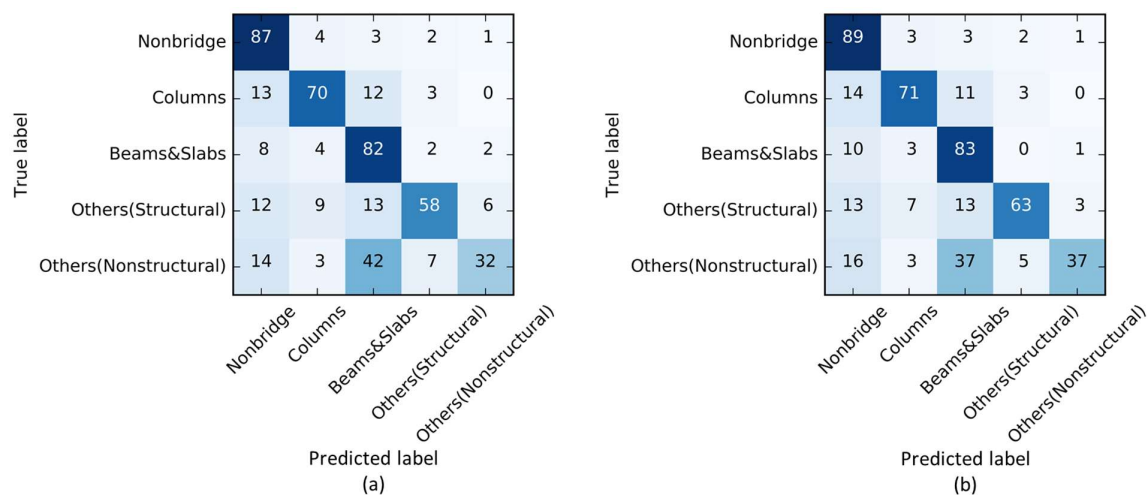

Figure 4 Confusion matrices of bridge component recognition
(a)With scene information (b) Without scene information

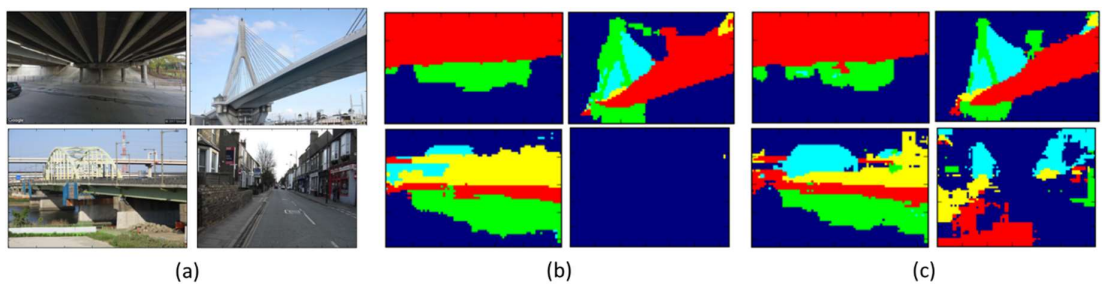

Figure 5 Example bridge component recognition results
Dark blue: Nonbridge, Green: Columns, Red: Beams&Slabs, Light blue: Other struct., Yellow: Other nonstruct.
(a)Input images (b)Recognition with scene understanding (c)Recognition without scene understanding

Table 5 False positive rates for the 9 scene classes with and without scene understanding

| Scene label | Bldg | Green | Person | Pvmt | S&P | Vhcls | Water | Sky | Other |
|---|---|---|---|---|---|---|---|---|---|
| FP (w/ scene) [%] | 1.8 | 0.6 | 0.1 | 0.6 | 0.3 | 0.2 | 2.3 | 0.1 | 3.3 |
| FP (w/o scene) [%] | 38.8 | 19.2 | 23.6 | 18.0 | 27.1 | 20.7 | 16.4 | 3.1 | 33.8 |

## 6. CONCLUSIONS

This paper proposed an approach to vision-based automated bridge component recognition to support the application of component-level damage detection methods to the images of complex scenes. The bridge component recognition in this study is based on the multi-scale convolutional neural networks (multi-scale CNNs), which are effective and compact CNN architectures capable of identifying objects photographed in different scales. To get the recognition results consistent with the high-level scene structures, a multi-scale CNN was trained to classify every image pixel into one of the 10 scene classes, i.e. Building, Greenery, Person, Pavement, Signs and Poles, Vehicles, Bridges, Water, Sky, and Others. Then, another multi-scale CNN for bridge component classification was trained on the images augmented by the softmax probabilities computed by the scene classifier. For the training and testing of the scene and bridge component classifiers, the labelled image data was collected by combining the existing datasets and the data labelled by the authors' research group.

For the scene classification, four network architectures inferred from the proven CNN architectures are trained and compared. The best network, *ResNet23*, achieved 88.68% of total pixel-wise accuracy, with 95% of the accuracy for bridges.

For the bridge component classification, the classifiers with and without scene understanding are trained and compared. By integrating the scene understanding, the total pixel-wise recognition accuracy decreased by 2.07%. However, when the classifiers are tested on non-bridge images, the approach with scene understanding showed significantly lower false-positive detections. Therefore, the bridge component recognition with scene understanding presented in this study works better than the naïve bridge component recognition for the images of complex scenes.

Although the bridge component recognition using multi-scale CNNs was successful in integrating the high-level scene structures into the recognition task, more research need to be done to improve the accuracy and consistency. Additional post-processing may be effective in imposing local or global consistency of labeling. Combining additional types of information, such as bridge structure types and coarse 3D information, may improve the accuracy. By combining an appropriate set of classifiers and additional processing, the automated bridge component recognition presented in this study is expected to contribute to the rapid initial post-earthquake assessment of bridges spread over the affected areas.

**ACKNOWLEDGEMENT**
We acknowledge the help of Mr. Peisong Wu, Mr, Guangpan Zhou, Ms. Xinxia Li, Mr. Yuguang Fu (the Smart Structures Technology Laboratory at the University of Illinois at Urbana-Champaign), Mr. Takahiro Yamaguchi, Mr. Mustafa Atta (the University of Tokyo) for labelling bridge photos. We also acknowledge Mr. Shintaro Arai for providing bridge photos used in this study.

**REFERENCES**

1. National Research Council (2011), National earthquake resilience: Research, implementation, and outreach, National Academies Press.
2. Ministry of Land, Infrastructure, Transport and Tourism (MLIT) (2012), 地域のモビリティ確保の知恵袋２０１２〜災害時も考慮した「転ばぬ先の杖」〜 [Lessons from the Great East Japan Earthquake, in Japanese]. Available: http://www.mlit.go.jp/sogoseisaku/soukou/sogoseisaku_soukou_tk_000037.html. [Accessed 12 6 2017].
3. East Japan Railway Company (JR-EAST), Ticket Exchange and Refund Policy. Available: https://www.jreast.co.jp/kippu/24.html. [Accessed 12 6 2017].
4. Ministry of Land, Infrastructure, Transport and Tourism (MLIT) (2012). 大規模地震発生時における首都圏鉄道の運転再開のあり方に関する協議会報告書 [A report of the committee on the resumption of train operation in Tokyo area after large-scale earthquakes, in Japanese]. Available: http://www.mlit.go.jp/common/000208774.pdf. [Accessed 13 6 2017].
5. Greer A. (2012). Earthquake Preparedness and Response: Comparison of the United States and Japan, *Leadership and Management in Engineering*. **12:3**, 111-125.
6. Yamazaki F., and Matsuoka M. (2007). Remote sensing technologies in post-disaster damage assessment. *Journal of Earthquake and Tsunami.* **1:03**, 193-210.

7. Galarreta F. J., Kerle N., and Gerke M. (2015). UAV-based urban structural damage assessment using object-based image analysis and semantic reasoning. *Natural Hazards and Earth System Sciences.* **15:6**, 1087-1101.
8. Xu Z., Yang J., Peng C., Wu Y., Jiang X., Li R., Zheng Y., Gao Y., Liu S., and Tian B. (2014). Development of an UAS for post-earthquake disaster surveying and its application in Ms7. 0 Lushan Earthquake, Sichuan, China. *Computers & Geosciences*. **68**, 22-30.
9. Zhu Z., German S., and Brilakis I. (2011). Visual retrieval of concrete crack properties for automated post-earthquake structural safety evaluation. *Automation in Construction.* **20:7**, 874-883.
10. Yeum C. M., and Dyke S. J. (2015). Vision-Based Automated Crack Detection for Bridge Inspection. *Computer‐Aided Civil and Infrastructure Engineering* **30.10**, 759-770.
11. Cha Y. J., Choi W., and Buyukozturk O. (2017). Deep learning-based crack damage detection using convolutional neural network. *Computer-Aided Civil and Infrastructure Engineering* **32:3,** 361-371.
12. German S., Brilakis I., and DesRoches R. (2012). Rapid entropy-based detection and properties measurement of concrete spalling with machine vision for post-earthquake safety assessments. *Advanced Engineering Informatics.* **26:4**, 846-858.
13. Yeum C. M., Dyke S. J., Ramirez J., and Benes B. (2016). Big visual data analytics for damage classification in civil engineering. *International Conference on Smart Infrastructure and Construction*. **Cambridge, UK**.
14. Zhu Z., and Brilakis I. (2010). Concrete Column Recognition in Images and Videos. *Journal of computing in civil engineering.* **24:6**, 478-487.
15. Farabet C., Couprie C., Najman L., and LeCun Y. (2013). Learning Hierarchical Features for Scene Labeling. *IEEE transactions on pattern analysis and machine intelligence.* **35:8**, 1915-1929.
16. Szeliski R. (2010) Computer vision: algorithms and applications, Springer Science & Business Media.
17. He K., Zhang X., Ren S., and Sun J. (2016). Deep Residual Learning for Image Recognition. *Proceedings of the IEEE Conference on Computer Vision and Pattern Recognition (CVPR 2016).* **Las Vegas, USA**.
18. Simonyan K., and Zisserman A. (2015). Very deep convolutional networks for large-scale image recognition. *Proceedings of International Conference on Learning Representations (ICLR 2015)*. **San Diego, USA**.
19. Krogh A., and Hertz J. A. (1991). A Simple Weight Decay Can Improve Generalization. *Advances in Neural Information Processing Systems (NIPS)*. **4**, 950-957.
20. Diederik K., and Ba J. (2014). Adam: A method for stochastic optimization. *Proceedings of International Conference for Learning Representations (ICLR 2015)*, **San Diego, USA**.
21. Srivastava N., Hinton G., Krizhevsky A., Sutskever I., and Salakhutdinov R. (2014) Dropout: A Simple Way to Prevent Neural Networks from Overfitting. *The Journal of Machine Learning Research.* **15.1**, 1929-1958.
22. Sergey I., and Szegedy C. (2015). Batch normalization: Accelerating deep network training by reducing internal covariate shift. *Proceeding of International Conference on Machine Learning (ICML 2015)*. **Lille, France**.
23. Eigen D., and Fergus R. (2015). Predicting Depth, Surface Normals and Semantic Labels with a Common Multi-Scale Convolutional Architecture. *Proceedings of the IEEE International Conference on Computer Vision (ICCV 2015)*. **Santiago, Chile**.
24. Abadi M., Barham P., Chen J., Chen Z., Davis A., Dean J., Devin M., Ghemawat S., Irving G., Isard M., Kudlur M., Levenberg J., Monga R., Moore S., Murray D. G., Steiner B., Tucker P., Vasudevan V., Warden P., Wicke M., Yu Y., and Zheng X. (2015). TensorFlow: Large-scale machine learning on heterogeneous systems. Available: https://www.usenix.org/system/files/conference/osdi16/osdi16-abadi.pdf. [Accessed 08 06 2017], Software available from tensorflow.org.
25. Krizhevsky A., Sutskever I., and Hinton G. E. (2012). Imagenet classification with deep convolutional neural networks. *Advances in neural information processing systems (NIPS)*, **25**, 1097-1105.
26. Krizhevsky A., and Hinton G. (2009). Learning multiple layers of features from tiny images. Technical Report, University of Toronto.
27. Gould S., Fulton R., and Koller D. (2009). Decomposing a Scene into Geometric and Semantically Consistent Regions. *Proceedings of International Conference on Computer Vision (ICCV 2009)*, **Kyoto, Japan**.
28. Liu C., Yuen J., Torralba A., Sivic J., and Freeman W. T. (2008). SIFT Flow: Dense Correspondence across Different Scenes. *European Conference on Computer vision (ECCV 2008)*, **Marseille, France**.
29. Ros G., Sellart L., Materzynska J., Vazquez D., and Lopez A. M. (2016). The SYNTHIA Dataset: A Large Collection of Synthetic Images for Semantic Segmentation of Urban Scenes. *Proceedings of the IEEE Conference on Computer Vision and Pattern Recognition (CVPR 2016)*. **Las Vegas, USA**.
30. Brostow G. J., Fauqueur J., and Cipolla R. (2009). Semantic object classes in video: A high-definition ground truth database. *Pattern Recognition Letters.* **30.2**, 88-97.
31. Brostow G. J., Shotton J., Fauqueur J., and Cipolla R. (2008). Segmentation and Recognition Using Structure from Motion Point Clouds. *European Conference on Computer vision (ECCV 2008)*, **Marseille, France**.
32. Google Street View. Available: https://www.google.com/streetview/
33. Google Street View Image API. Available: https://developers.google.com/maps/documentation/streetview/